\documentclass[letterpaper, 10 pt, conference]{ieeeconf}  

\IEEEoverridecommandlockouts                              

\usepackage{multicol}
\usepackage[bookmarks=true]{hyperref}
\usepackage{tabularx}
\usepackage{tabulary}
\usepackage{array}
\usepackage{multirow}
\usepackage{url}            
\usepackage{graphicx} 
\usepackage[dvipsnames]{xcolor}

\usepackage{cuted}
\usepackage{gensymb}
\usepackage{multirow}
\usepackage{caption}
\usepackage[noadjust]{cite}

\hypersetup{
    colorlinks=true,
    linkcolor=blue,
    filecolor=magenta,      
    urlcolor=cyan,
}

\newcommand{\name}{{PolyTouch}}
\newcommand{\website}{https://polytouch.alanz.info/}

\newcommand\best[1]{\textcolor{Green}{#1}}
\newcommand\worst[1]{\textcolor{Maroon}{#1}}
\newcolumntype{P}[1]{>{\centering\arraybackslash}p{#1}}

\title{\LARGE \bf
\name{}: A Robust Multi-Modal Tactile Sensor for Contact-rich Manipulation Using Tactile-Diffusion Policies}

\author{Jialiang Zhao$^{1}$, Naveen Kuppuswamy$^{2}$, Siyuan Feng$^{2}$, Benjamin Burchfiel$^{2}$, Edward Adelson$^{1}$%
\thanks{$^{1}$ MIT CSAIL,%
        {\tt\small \{alanzhao, adelson\}@csail.mit.edu}}%
\thanks{$^{2}$Toyota Research Institute,
        {\tt\small firstname.lastname@tri.global}}%
\\
This paper has been nominated for the \textcolor{red}{Best Paper Award} at ICRA 2025.
}

\begin{document}

\maketitle
\thispagestyle{empty}
\pagestyle{empty}

\vspace{15pt}  


\begin{strip}
\vspace{-5em}
\centering
\includegraphics[width=\linewidth]{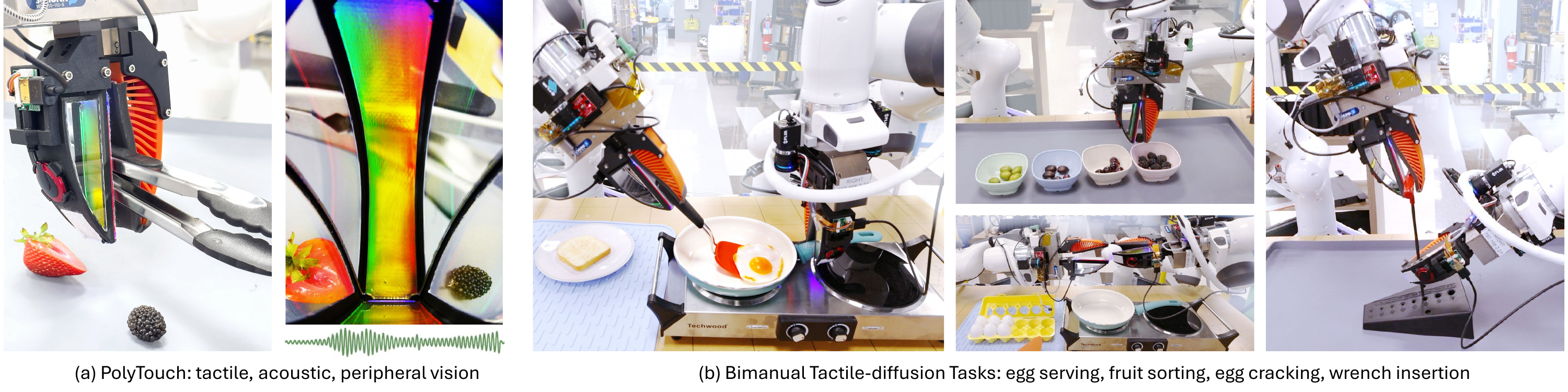}
\captionof{figure}{
(a) \name{} is a robot finger that combines tactile, acoustic, and peripheral vision sensing. 
(b) We design 4 common bimanual tasks to evaluate tactile-diffusion policies: egg serving, fruit sorting, egg cracking, and wrench insertion. 
}
\label{fig:banner}
\end{strip}

\begin{abstract}
Achieving robust dexterous manipulation in unstructured domestic environments remains a significant challenge in robotics. Even with state-of-the-art robot learning methods, haptic-oblivious control strategies (i.e. those relying only on external vision and/or proprioception) often fall short due to occlusions, visual complexities, and the need for precise contact interaction control. To address these limitations, we introduce \name{}, a novel robot finger that integrates camera-based tactile sensing, acoustic sensing, and peripheral visual sensing into a single design that is compact and durable. \name{} provides high-resolution tactile feedback across multiple temporal scales, which is essential for efficiently learning complex manipulation tasks. Experiments demonstrate an at least 20-fold increase in lifespan over commercial tactile sensors, with a design that is both easy to manufacture and scalable. We then use this multi-modal tactile feedback along with visuo-proprioceptive observations to synthesize a tactile-diffusion policy from human demonstrations; the resulting contact-aware control policy significantly outperforms haptic-oblivious policies in multiple contact-aware manipulation policies. This paper highlights how effectively integrating multi-modal contact sensing can hasten the development of effective contact-aware manipulation policies, paving the way for more reliable and versatile domestic robots.
More information can be found at \url{\website}.

\end{abstract}

\section{Introduction}

Achieving robust and reliable dexterous manipulation in the real world is a hard open challenge, particularly in the case of unstructured domestic environments. This challenge is further exacerbated by the desire to have domestic robots capable of diverse range of skills. 
Increasingly, we look to machine learning methods to provide feasible solutions. In particular, policy synthesis through supervised behavior learning from human demonstrations is emerging as a powerful alternative to traditional methods in achieving the necessary skill diversity~\cite{team2024octo,wang2024hpt,chi2023diffusion}. 
A key element of the challenge is in coping with uncertain contact interactions.
Most skills are inherently contact-rich, i.e. they typically involve multiple contact interactions throughout the skill. 



A key factor to consider is that haptic-oblivious control policies, i.e. ones that use only external vision and/or propioception as feedback may be fundamentally ill-equipped to deal with the inherent complexity. 
Challenging factors are often encountered in various skills, e.g. visually complex scenes (cluttering), external / self-occlusion, objects or environment being hard-to-see (transparent, reflective) or hard-to-manipulate (articulated, deformable). Moreover, successful skill execution may require precise regulation of contact forces or impedance while coping with external disturbances.

In this context, incorporation of haptic feedback within control policy frameworks presents an important and viable approach to alleviating the challenges; tactile sensors present a direct way to sense contact-state and therefore act as an additional information source for inferring state of the manipuland and the environment~\cite{bronars2024texterity,zhao2023fingerslam,suresh2023neural}. However, two key questions need to be answered: (a) conceptually, what is the right kind of sensor modality and the required signal processing architecture? and (b) practically, what is the optimal design for a compact, robust, and easy-to-build sensor for large-scale data-driven policy synthesis? 

A recent development in sensor design is that of camera based tactile-sensing, i.e. coupling a camera with a deformable reflective membrane in order to capture the contact interaction~\cite{yuan2017gelsight}. 
They offer a high-resolution solution for fine texture-based tactile sensing, which recent literature has demonstrated to be crucial for manipulation tasks requiring precise control and heightened awareness of shape or object characteristics~\cite{burgess2024learning,yuan2015measurement,dong2017improved}.
Contact microphone-based vibrational sensing is another tactile sensing solution that captures the sound wave generated during contact with an object. 
Analysis of the sensed vibrations can reveal properties like texture, hardness, and contact events~\cite{liu2024maniwav,jin2019open}. 
In addition, proximity sensing through peripheral vision has also been shown to be useful for dexterous manipulation in clutter \cite{yamaguchi2016combining}. 
Inspired by some of these results, we seek a design that combines all of these modalities, leveraging the high-frequency capabilities of vibrational sensing, the high spatial resolution of camera-based tactile sensing, and the utility of peripheral sensing in clutter. 

\begin{figure}[h!]
\centering
\includegraphics[width=\linewidth]{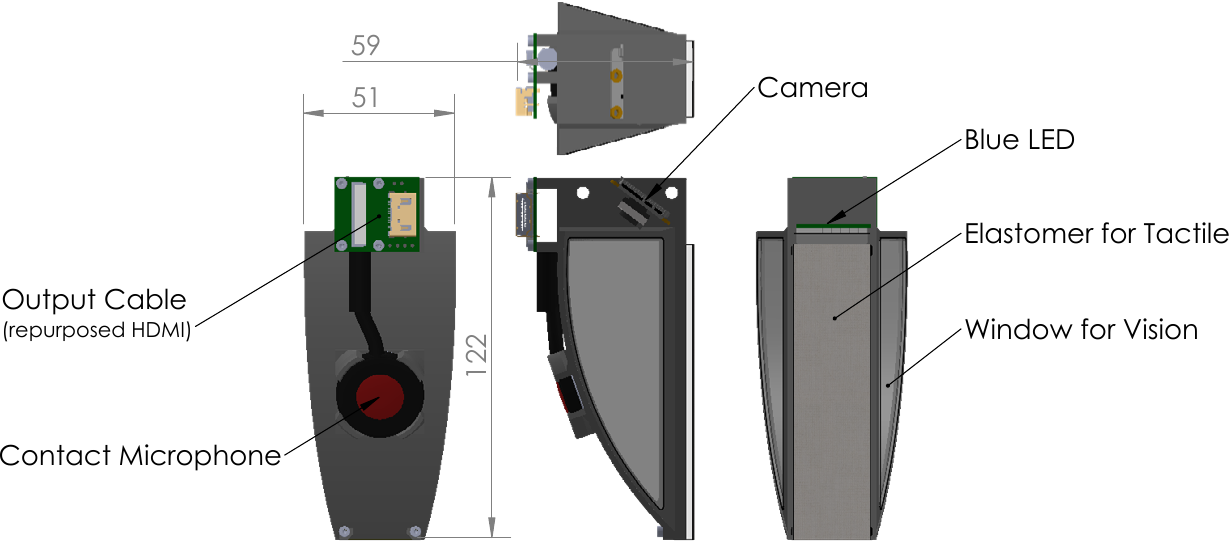}
\caption{\textbf{Drawings and main components of \name{}}}
\label{fig:threeviews}
\end{figure}

This paper presents \name{}, a novel robot finger that combines camera-based texture sensing, acoustic sensing, and peripheral visual sensing into one robust, compact, and easy-to-manufacture design. The three modalities incorporated into the finger are essential for robots to efficiently learn complex and contact-rich manipulation policies. Furthermore, it is designed to address several critical issues of contemporary tactile sensors that hinder their scalability in large data domains, such as poor durability and manufacturability. Our experiments show that \name{} has a lifespan at least 20 times longer than state-of-the-art sensors. Additionally, its fabrication process does not require specialized equipment or expertise, making it an ideal sensor for large-scale, on-robot policy synthesis. Specifications of \name{} are listed on Tab.~\ref{tab:spec} and Fig.~\ref{fig:threeviews}.

This paper then leverages this sensor for the synthesis of contact-aware manipulation policies using multi-modal sensing. We build upon the diffusion policy to develop an architecture that exploits cross-modal attention. This tactile-diffusion policy framework is then used to characterize and demonstrate the benefits of incorporating haptic feedback for contact-rich manipulation.

\begin{table}[!h]
  \caption{\name{} Specifications}
  \label{tab:spec}
  \centering
  \begin{tabular}{m{20mm}|m{55mm}}
    \hline
    Elastomer Replacement & Gel cartridge can be quickly swapped out and replaced with new ones. \\
    \hline
    Elastomer Options & VHB tape based option (acrylic foam base material, semi-specular paint) and silicone option (silicone rubber base material, lambertian paint) \\
    \hline
    Durability & $>$35hrs under continuous tool using\\
    \hline
    Modalities & Texture, acoustic, and peripheral vision \\
    \hline
    Output Format & One Ethernet output (video and audio streams) \\
    \hline
    Dimension & 51mm (L) x 59mm (W) x 122mm (H) \\
    \hline
    FoV for tactile & 100mm x 25mm \\
    \hline
    
  \end{tabular}
\end{table}

\section{Related Works}
\label{sec:related_works}


\textbf{Camera-based texture sensing: high spatial resolution for delicate manipulation.}
Camera-based tactile sensors, such as the GelSight Family~\cite{yuan2017gelsight}, DenseTact~\cite{do2022densetact}, Bubble Grippers~\cite{bubblegrippers,bigbubble}, GelSlim~\cite{donlon2018gelslim} produce high resolution video streams of the texture of the contact surface. 
They generally incorporate a soft contact medium (silicone gel in GelSight, DenseTact, and GelSlim, inflated air in Bubble) that conforms to the shape of the external object when in contact, and a camera sitting on the opposite side captures high resolution images of the textures and shapes of the contact.
However, despite the highly detailed information that those sensors provide, several well-known limitations exist that pose a challenge to their widespread adoption.
\textbf{Resolution v.s. speed}: The high spatial resolution comes at the trade-off of lower frequency due to camera frame rates (typically below 100 Hz).
Dynamic information on contact, such as impacts and vibrations due to slippage, requires sensing at much higher rates.
\textbf{Poor durability and manufacturability}: the soft elastomer endures constant abrasion and torsion. Scratches, tearing, and delamination are common issues especially for silicone-based sensors. The construction of the soft elastomer often requires specialized equipment and expertise, which hinders their wide adoption.
\textbf{Bulkiness:} a larger sensing surface typically necessitates positioning the camera further from the contact surface due to optical and illumination requirements, which consequently adds bulk to the system.

\textbf{Contact microphone-based vibration sensing: high temporal frequency for dynamic manipulation.}
A number of research have shown that the higher frequency vibrational signals captured by contact microphones could lead to better manipulation performances in manipulation tasks that require timely impact type and collision detection such as in food cutting~\cite{zhang2019leveraging}, human-robot interaction~\cite{fan2020acoustic,fan2021aurasense,gamboa2020detecting}, and force level estimation~\cite{ono2015sensing}.
Liu et al. installed a contact microphone on a hand-held robot gripper to collect in-the-wild data, which was then used to successfully train robotic policies and proved the usefulness of acoustic data in improving manipulation quality for a range of dynamic tasks~\cite{liu2024maniwav}.

\textbf{Policy learning with multi-modal sensing.}
Incorporating multi-modal sensing into robot learning has been an active area of research.
Diffusion policy \cite{chi2023diffusion}, a recently proposed method for visuomotor policy learning has been proven very effective and versatile in generating complex multi-modal actions. 
The approach uses a conditional denoising diffusion process on the action-space of the robot, resulting in a policy formulation that can express arbitrary multi-modal action distributions, capture high-dimensional action spaces, and is stable in training. 
A number of works have chosen diffusion policy as the action prediction head under different contexts, such as cross-embodiment learning~\cite{team2024octo} and multi-modal sensing~\cite{wang2024poco}. 

The proposed design aims to combine texture, vibration, visual sensing and at the same time overcomes the reliability and manufacturability issues that contemporary tactile sensors suffer from.
We also propose a tactile-diffusion framework extended from diffusion policy that combines multi-modal sensing in a coherent manner, and we show that the additional modalities improve manipulation performance.

\section{Mechanical Design and Specifications}
\label{sec:design}


The design goal of \name{} is threefold: the finger needs to be sensitive, durable, and easy to manufacture.

\subsection{Sensing}
\name{} is equipped with three sensing modalities: camera-based tactile sensing, acoustic sensing, and peripheral natural vision sensing.

\begin{figure}[h!]
\centering
\includegraphics[width=\linewidth]{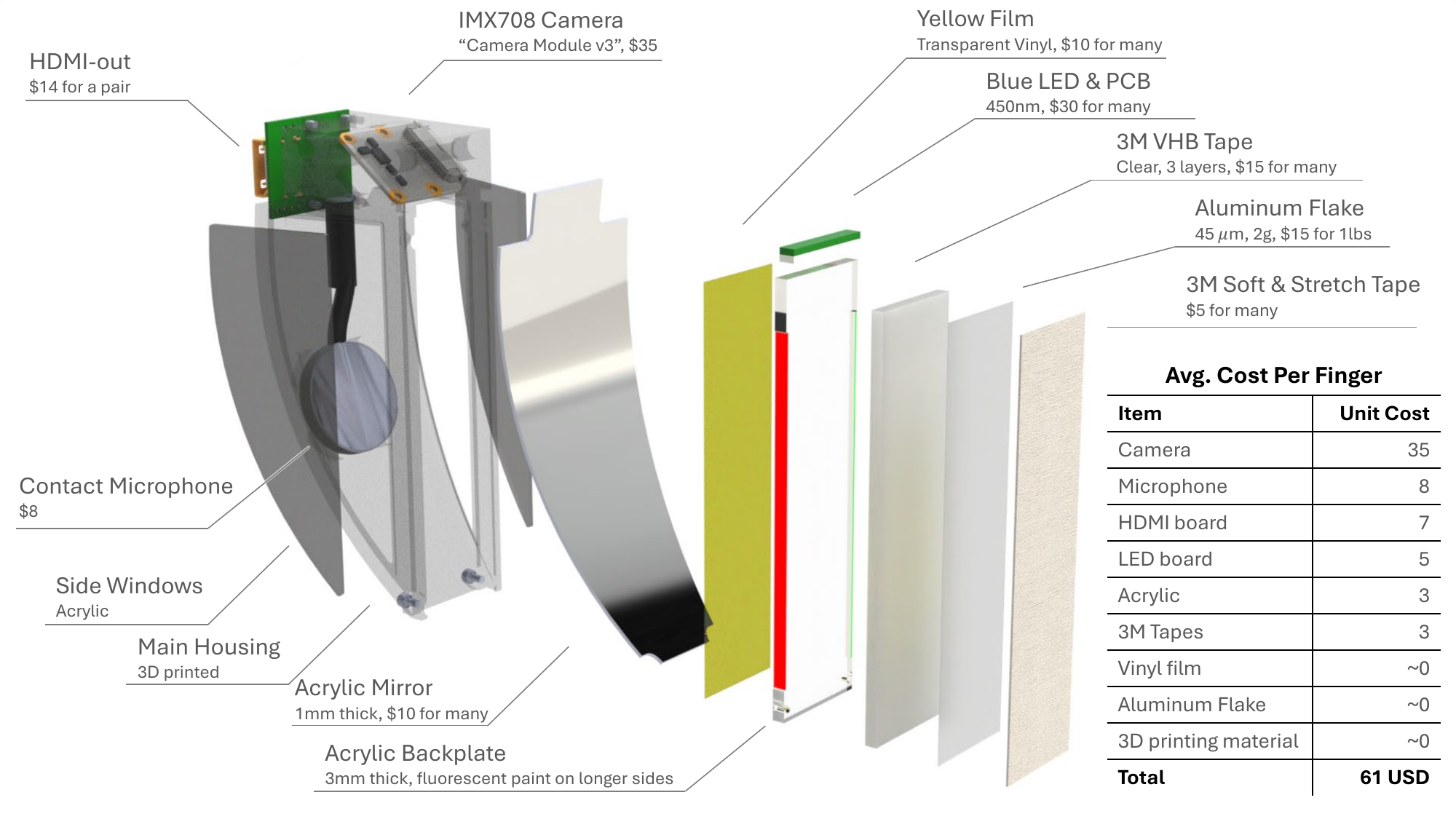}
\caption{\textbf{Explosion view of \name{} and its estimated cost.} \name{}'s BoM is mostly comprised of easily accessible materials, and its construction does not require specialized equipment.}
\label{fig:material}
\end{figure}

\subsubsection{\textbf{Camera-based tactile sensing}}
In \name{} tactile sensing is achieved by embedding a RGB camera inside the finger pointing at a clear elastomer.
We provide two exchangeable options as the clear elastomer:
\begin{itemize}
    \item \textbf{VHB Elastomer}: Clear \textit{3M VHB} double-sided tape, with its out-facing surface covered with $45\mu m$ reflective aluminum powder.
    \item \textbf{Silicone Elastomer}: \textit{Silicones Inc.} XP-565 silicone rubber with \textit{PRINT-ON} gray silicone ink on its out-facing side.
\end{itemize}
The VHB elastomer is much easier and faster to construct than the silicone elastomer.
However, the VHB tape is a viscoelastic material that takes some time to recover to its original shape after a contact is removed, while the silicone elastomer responds much faster.
The elastomer sits on top of a clear acrylic back plate.
The bottom shorter side of the acrylic is illuminated with blue LEDs ($\lambda = 450nm$).
The two longer sides are painted with pink and green fluorescent paint (Liquitex BASICS, Fluorescent Pink and Fluorescent Green).
The two colors of fluorescent paint absorbs the blue light which has a shorter wavelength, then emit the energy as longer wavelength light in pink and green.
The method of using fluorescent paint as illumination sources in tactile sensors was first introduced by Adelson in~\cite{adelson2024retrographic}, and it has the advantage of reduced bulkiness and power consumption compared to using multi-colored LEDs as in GelSight, DIGIT, and similar sensors~\cite{yuan2017gelsight,lambeta2020digit}.
The pink and green lights emitted by the fluorescent paint are weaker than the blue light emitted by the LED.
Although human eyes can distinguish the unbalanced colors with ease, the blue light can easily saturate most optical sensors.
To solve this issue, a transparent yellow vinyl filter is glued to the inner side of the acrylic back plate to reduce the amount of blue light that reaches the camera.

\begin{figure}[h!]
\centering
\includegraphics[width=\linewidth]{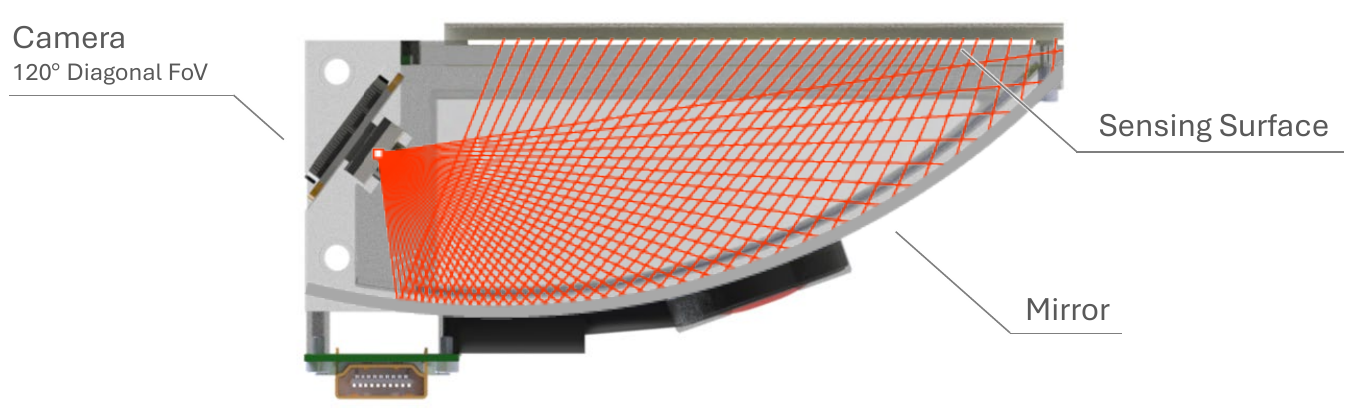}
\caption{\textbf{Optical simulation of \name{}}. The light reflected by the sensing surface is redistributed by the curved mirror before reaching the camera. The camera sensor is 4:3 with a diagonal field of view of 120 degrees.}
\label{fig:ray_tracing}
\end{figure}

To achieve a long and large sensing area without needing more than one cameras or increasing the size of the finger, a curved mirror is placed at the back of the finger, similar to GelSight Svelte~\cite{zhao2023gelsight,zhao2023sveltehand} and GelLink~\cite{ma2024gellink}.
The curvature and placement of the mirror is designed so that 
(1) the entire inner surface of the acrylic back plate is visible from the camera's field of view;
(2) the camera should have a near-orthogonal view of the back plate, so that light from all directions are visible to the camera which means the depth information embedded in the colors will be better preserved;
and (3) the distortion introduced by the curved mirror should be minimized.
A reflection simulation result of the final design is illustrated in Fig.~\ref{fig:ray_tracing}.

\subsubsection{\textbf{Contact microphone-based acoustic sensing}}
A piezoelectric contact microphone is placed on the back of \name{} to gather acoustic signals.
The contact microphone samples at 48 kHz and the audio file is saved or streamed in synchronization with the tactile video stream.
Both audio and video are recorded by a Raspberry Pi, which then transcodes the signals before saving them as .mp4 files or streaming to robot stations.

\subsubsection{\textbf{Peripheral vision sensing}}
The camera inside \name{} not only gathers tactile information but also peripheral visual signals with the help of two side windows.
The side windows together with the mirror help achieve a visible area of both the surrounding areas of the contact surface and the area underneath the sensor.
Demonstration of the visible areas of peripheral visual sensing are illustrated in Fig.~\ref{fig:banner}.

\subsection{Durability and replaceability}
The durability issues with camera-based tactile sensors are usually associated with the elastomers found in those sensors, which have to be in direct and repetitive contact with the outer environment.
The issues are twofold: the elastomers can delaminate under excessive force/torque and the surface wears or tears under continuous abrasion.  
The delamination problem mostly affects sensors using silicone gel as elastomers, due to the difficulty in gluing silicone to non-silicone materials while maintaining optical clarity.
The VHB Elastomer option of \name{} provides an alternative where itself is highly adhesive, which eliminates the delamination problem almost entirely. 

Abrasion and torsion resistance can be improved by applying protective films onto the sensing surface of the sensors.
One popular such choice is 3M Tegaderm films, which were originally introduced as a protection film for wounds.
However, this film is thin with limited lifespan against constant abrasion and torsion, and it wrinkles easily.
Instead, \name{} chooses 3M Nextcare Soft \& Stretch tape as the outer protection layer, which has a similar surface property with human skins, and it does not wrinkle easily. 

\name{} is also designed with a quick elastomer swapping mechanism so that the gel could be easily and quickly slided out and replaced with a new one.

\subsection{Manufacturability}
\name{} is designed to minimize the requirement of special equipment or expertise in construction.
The fabrication of the elastomer of \name{}-VHB does not require any specialized equipment or expertise: it simply requires applying the tape onto the laser cut acrylic back plate, then wiping aluminum powder on the outer side of the tape. 
Making one VHB elastomer takes less than 5 minutes for an unexperienced person with basic hands-on skills.
The fabrication process of the silicone gel in \name{}-Silicone is silimar to other popular soft sensors, e.g. GelSight~\cite{yuan2017gelsight}, DenseTact~\cite{do2022densetact}, ReSkin~\cite{bhirangi2021reskin}.

\section{Robot Learning from Multi-Modal Sensing}
\label{sec:nn}
We employ a fixed-base bi-manual hardware platform that consists of two Franka Panda robot arms mounted in an upright configuration as shown in Fig.~\ref{fig:banner}. 
Each arm is equipped with one wrist-mounted camera (FLIR Blackfly S).
Two fixed scene cameras (FRAMOS D415e) are positioned to have good visibility over the active workspace. 
The robots themselves are equipped with 2 kinds of fingers on each gripper: a grooved passive compliant \emph{fin-ray} \cite{crooks2016fin} 3D printed with TPU and a \name{}-VHB finger.
This particular gripper configuration affords good grasps on arbitrarily shaped manipulands as well as for maintaining a strong grasp on tool handles.
All of the skills that we analyze in this paper are demonstrated on a workbench adjacent to the robot setup, and the robots are tele-operated with two 6-DoF spacemouse during data collection.

\begin{figure*}[t!]
\centering
\includegraphics[width=0.8\linewidth]{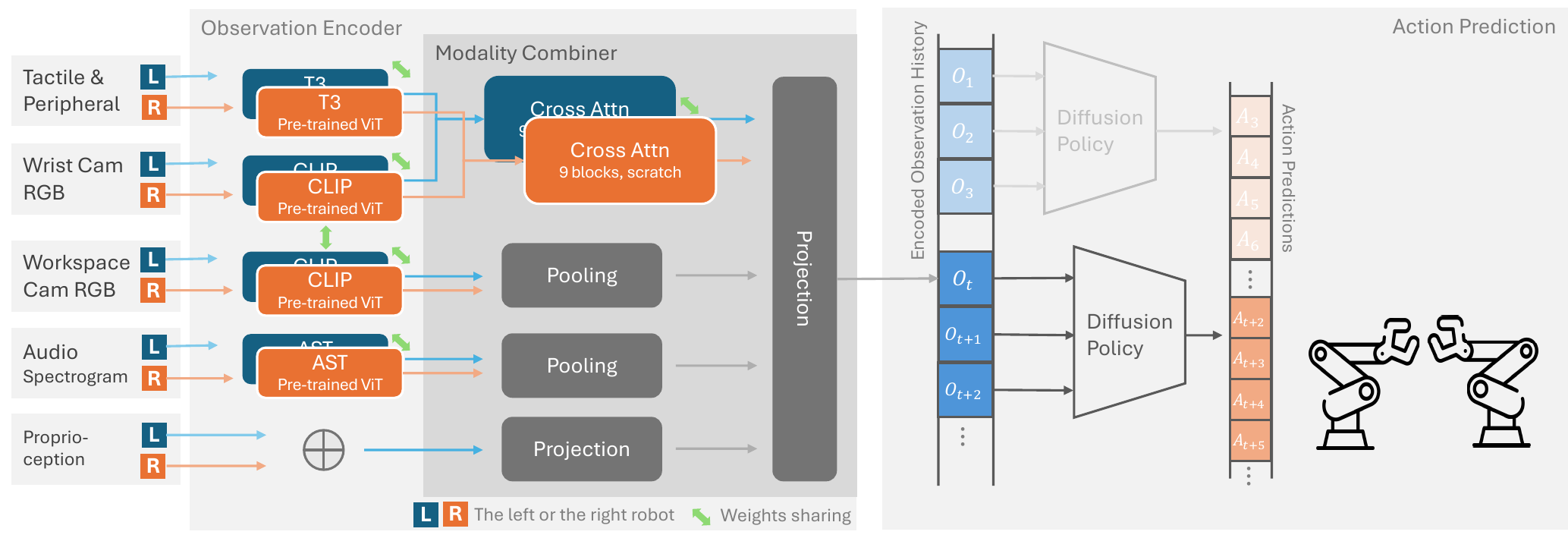}
\caption{\textbf{The tactile-diffusion policy network.} All sensing modalities coming from each of the two arms are encoded with pre-trained feature extractors (except proprioception which is encoded by a MLP from scratch), then combined by a mixture of cross attentions and concatenations, before passed into a diffusion head for action prediction.}
\label{fig:nn}
\end{figure*}

\subsection{Modality Encoding}
The observation at time step $t$ and their encoding method is detailed as follows:

\begin{itemize}
    \item Tactile and peripheral vision $O_{tp}^t$ are obtained as RGB images from the \name{} mounted on each arm. They are encoded by a pre-trained T3~\cite{zhao2024transferable} encoder. T3 is a tactile representation learning framework that was pre-trained from a large scale multi-sensor and multi-task dataset.
    \item Wrist view $O_{wrist}^t$ and scene view $O_{scene}^t$ are obtained as RGB images from both arms. They are encoded by a pre-trained CLIP feature extractor~\cite{radford2021learning}.
    \item Audio signals $O_{aud}^t$ are obtained as sound waves from the \name{} mounted on each arm. They are first transformed to log-mel spectrogram then passed into an Audio Spectrogram Transformer (AST,~\cite{gong2021ast}) for feature extraction.
    \item Proprioception $O_{prop}^t$ for both arms, including actual and desired 6D EE poses and gripper widths, are encoded by a MLP before combined with other modalities.
\end{itemize}

A \textit{Modality Combiner} is designed to combine the encoded features from each modality.
The backbone architecture of the pre-trained CLIP vision feature extractor and the pre-trained T3 tactile feature extractor are both Vision Transformers~\cite{dosovitskiy2020image}, and the encoded features from $O_{tp}^t$ and $O_{scene}^t$ are combined with a 6-block 12-head cross attention.
The output is then pooled (by extracting the classification token) and combined with pooled or projected features from other modalities with a concatenation and projection layer.

\subsection{Policy Learning}
The combined features, denoted as $O^t$, is fed to a diffusion policy (U-Net variation) backbone for action prediction (observation history: 2, action prediction horizon: 16, action to execute: 8).

\section{Experiments and Discussions}
\label{sec:exp}


A durability experiment is designed to test \name{}'s lifespan, and a manipulation learning experiment is designed to test the usefulness of \name{}'s additional modalities.

\subsection{Durability Testing}
\begin{figure}[h!]
\centering
\includegraphics[width=\linewidth]{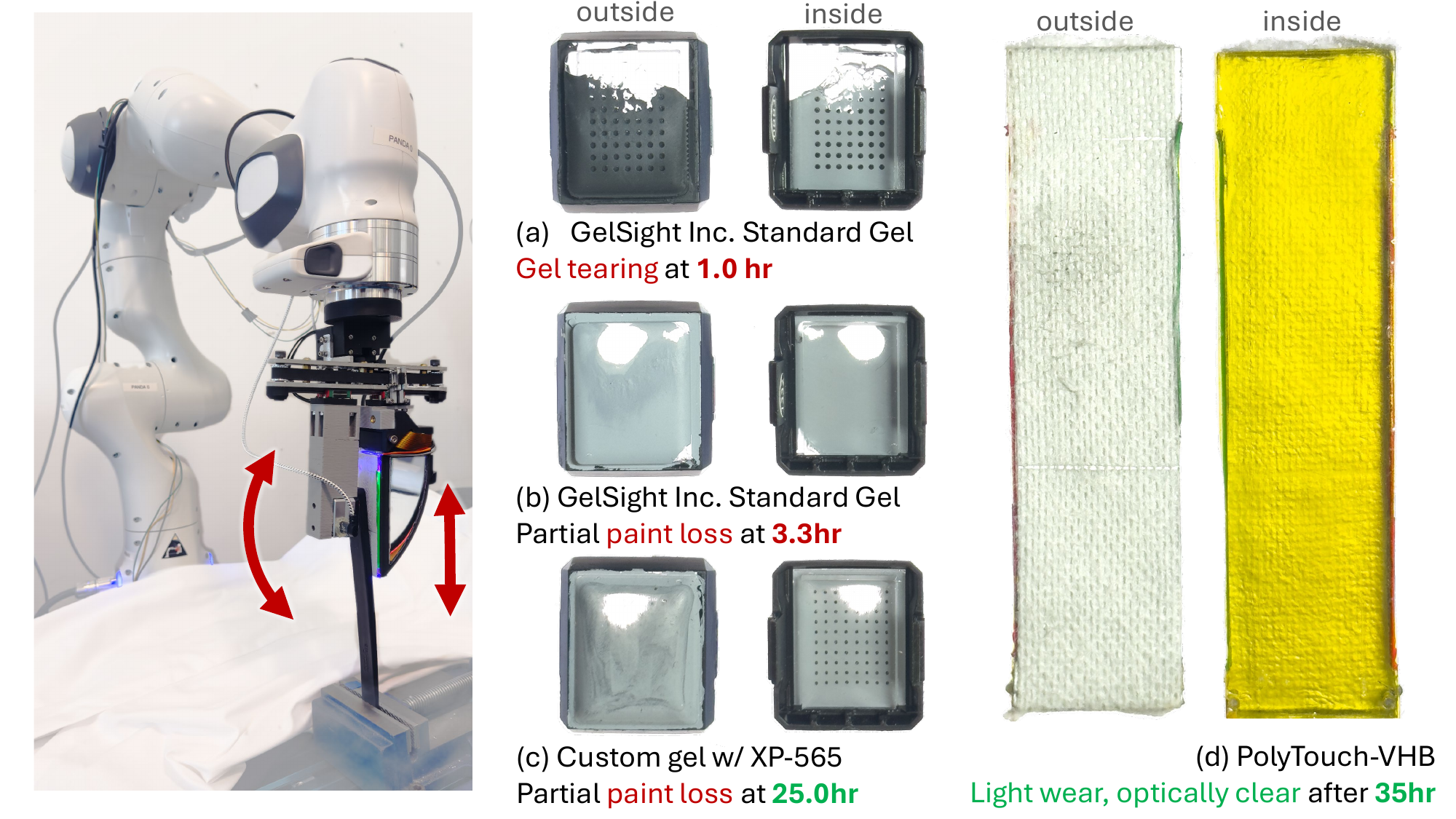}
\caption{\textbf{Elastomer durability testing under an emulated tool-using setting}. 
A Franka Panda robot performs continuous rubbing and chafing on a fixed flexible spatula handle.
A GelSight Mini made by GelSight Inc. and a \name{}-VHB are mounted opposing each other.
}
\label{fig:durability}
\end{figure}

We compare \name{}'s lifespan against that of GelSight Mini~\cite{gsmini}, a popular and commercially available camera-based tactile sensor.
Our experiment emulates the abrasion and torsion endured by robot fingers in household tool-using tasks.
The robot gripper is equipped with a GelSight Mini sensor on an extended finger on one side, and \name{}-VHB on the other side.
A plastic spatula is mounted at a fixed location on the workbench.
The robot grasps the handle of the spatula with a force randomly sampled from 10 to 30 N.
The robot then applies continuous random 6D movements, with translation sampled from -10 to 10 mm and rotation sampled from -5 to 5 degrees on each of x, y, z axes. 
Time before failures and pictures of the sensor surfaces are shown in Fig.~\ref{fig:durability}.

\name{}-VHB endured in total of 35hrs of continuous rubbing without failure or decreased image quality.
Two different standard gels from GelSight Inc. available for GelSight Mini were tested, where one last 1.0hr before complete gel separation while the other last 3.3hrs before paint loss.
Our updated silicone gel formula with XP-565 (the same material as on \name{}-Silicone) last 25.0hrs on GelSight Mini.

\subsection{Multi-modal sensing for bimanual manipulation}

\begin{table*}[ht!]
\begin{center}
\begin{tabular}{P{3cm}|P{10mm}|P{10mm}|P{10mm}|P{15mm}|P{10mm}|P{10mm}|P{10mm}|P{15mm}}
\hline
\multirow{2}{*}{\textbf{Task}} & \multicolumn{4}{c|}{\textbf{Avg. Task Progress}} & \multicolumn{4}{c}{\textbf{Avg. Task Success}} \\
\cline{2-9}
\rule{0pt}{3ex} & \textit{\textbf{visuo-\newline proprio}}  & \textit{\textbf{multi-\newline concate}} & \textit{\textbf{multi-\newline crossatn}} & Absolute improvement & \textit{\textbf{visuo-\newline proprio}} & \textit{\textbf{multi-\newline concate}} & \textit{\textbf{multi-\newline crossatn}} & Absolute improvement \\
\hline
\rule{0pt}{3ex}
Insert Wrench & \worst{47\%} & \best{60\%} & 59\% & +12\%          & \worst{0\%} & \best{20\%} & 18\% & +18\%\\
Sort Fruit & \worst{53\%} & 63\% & \best{70\%} & +17\%             & \worst{33\%} & 46\% & 46\% & +13\%\\
Crack Egg & \worst{70\%} & 71\% & \best{72\%} & +1\%              & \worst{50\%} & 53\% & \best{54\%} & +3\%\\
Serve Egg (1/3 data) & \worst{10\%} & - & \best{17\%} & +7\%      & 0\% & - & 0\% & -\\
Serve Egg (2/3 data) & \worst{53\%} & - & \best{56\%} & +3\%      & \worst{13\%} & - & \best{33\%} & +20\%\\
Serve Egg (all data) & \worst{81\%} & 88\% & \best{100\%} & +19\%  & \worst{66\%} & 73\% & \best{100\%} & +34\%\\
\hline
\end{tabular}
\end{center}
\caption{Evaluation performance on each task and network variant. Best performance is marked in \best{green} and worst in \worst{red}. Absolute improvement is the difference between \textbf{\textit{multi-crossatn}} (proposed method) and \textbf{\textit{visuo-proprio}} (baseline).}
\label{tab:eval_results}
\end{table*}

We design 4 tasks and 3 network variants for this section:

\textbf{BimanualInsertWrench}: one robot picks up a T-handle Allen wrench, passes on to another robot, the two robots then collaborate to aim and insert the wrench until fully inserted.
Around 200 datapoints are collected for training.

\textbf{BimanualSortFruit}: four fruits of similar visual appearance but different surface texture and softness are sorted into bins by two arms.
Around 150 datapoints are collected for training.

\textbf{BimanualEggCracking}: a two-piece toy egg is picked up by one arm, then split open above a pan with the help of the other arm.
Around 70 datapoints are collected for training.

\textbf{BimanualEggServing}: one robot picks up a spatula with the help of the other arm, then the robot scoops under a toy pan-fried egg with the spatula, finally the egg should be gently placed on top of a slice of toast.
Around 150 datapoints are collected for training.

The amounts of datapoints to collect for each task were chosen empirically based on task difficulty.
Illustration of each task is shown in Fig.~\ref{fig:banner}.
We train and evaluate three network configurations for each task: 

\textbf{\textit{visuo-proprio}} is the variant where the tactile-peripheral vision modality $O_{tp}^t$ and the audio signal $O_{aud}^t$ are masked out (replaced with zeros), and no cross attention is used (the encoded output is directly pooled and projected);

\textbf{\textit{multi-concate}} uses all modalities, and directly pooling and projection are used in place of cross attentions;

\textbf{\textit{multi-crossatn}} is the proposed method as shown in Fig.~\ref{fig:nn}, where all modalities are utilized together with the cross attention module.

Other than whether to use cross attention and whether to mask out tactile inputs, all network variants share the same architecture and hyper-parameters.
For the \textbf{BimanualEggServing} task, we added two more experiments to train with $1/3$ and $2/3$ of the total data.
All experiments were trained for 500 epochs with a batch size of 10 on AWS G5.48xLarge nodes ($8\times$ Nvidia A10G each).
During both training and evaluation, randomization on the object / tool init poses was performed.
We report evaluation results on two metrics: \textit{average task progress} and \textit{average task success}.
Each task is split into 3-7 stages.
The \textit{average task progress} refers to the mean progress of stages that each evaluation run successfully completes, while the \textit{average task success} represents the averaged binary success rate for each evaluation run.

We emphasize that (1) the \textbf{\textit{visuo-proprio}} policy is a state-of-the-art visuomotor policy coupled with an ample amount of video feeds (4 in total); (2) the four tasks evaluated are standard tasks that were not specifically curated for tactile sensing, unlike those commonly found in many previous tactile manipulation studies.
Our experiments shows that:
\subsubsection{Tactile-inclusive policies are more robust than SOTA visuomotor policy}
Policy variants that use tactile sensing modalities as inputs consistently outperformed the \textbf{\textit{visuo-proprio}} policy, as shown in Tab.~\ref{tab:eval_results}.

\begin{figure}[h!]
\centering
\includegraphics[width=0.9\linewidth]{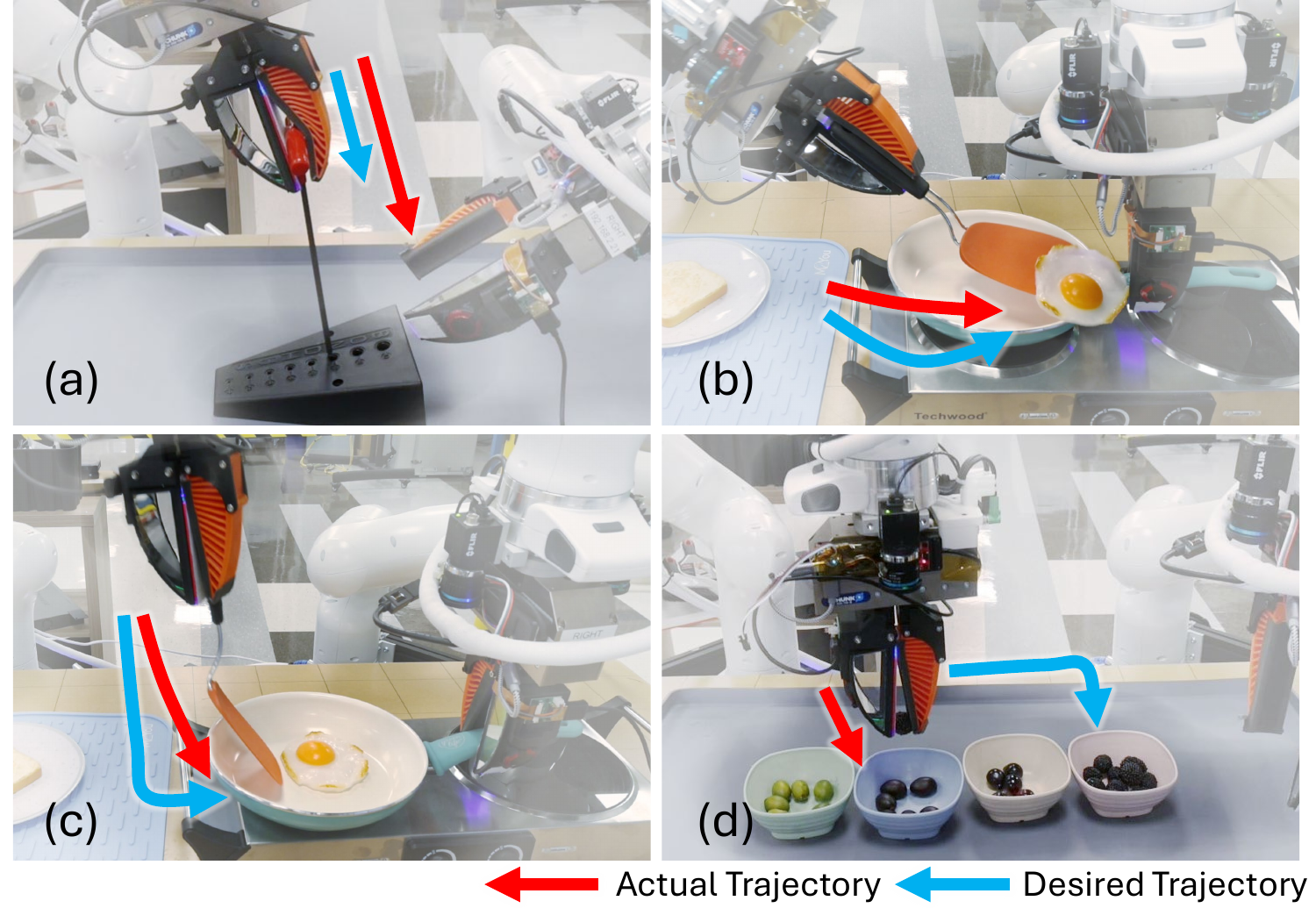}
\caption{\textbf{Failure modes unique to the \textbf{\textit{visuo-proprio}} policy}. 
(a) Excessive force applied in wrench insertion.
(b) Insufficient angle when scooping under a pan-fried egg.
(c) Insufficient force to press down spatula during a scooping motion.
(d) Wrongly sorted fruits with subtle visual differences.}
\label{fig:failures}
\end{figure}

\subsubsection{Some failure modes are unique to tactile-oblivious policies} Those failure modes did not or rarely appeared in either of the tactile-inclusive policies.

\textbf{Excessive or insufficient force} was observed in two scenarios, including applying too much force when inserting a wrench and not pressing the spatula hard enough when scooping the egg.
    These two issues did not happen at all with either of the tactile-inclusive policies.
    We believe tactile sensing, especially the texture-based sensing modality provides information to the policy learner to better modulate force.

\textbf{Less precision in grasping} is observed more frequently in the \textbf{\textit{visuo-proprio}} policy (11 instances during evaluation) than in tactile-inclusive policies (3 instances on average), including grasping the tool handle and the fruit too close to the edge.
    We hypothesize that the peripheral vision modality contributes to reducing such precision issues because it provides a close up view both right before and during a grasping process.

\textbf{Texture-based classification} in the fruit sorting experiment, especially the blackberry v.s. blueberry sorting task, shows a much higher success rate with tactile-inclusive policies (80\% v.s. 20\% with \textbf{\textit{visuo-proprio}}).
    This demonstrates that tactile could be pivotal in distinguishing visually similar objects based on surface textures.

\subsubsection{Training robotic policies with more modalities might require more data}
We notice a decrease in absolute performance improvement when we train a policy with less data.
Specifically, the average task progress improved 19\% with the \textbf{\textit{multi-crossatn}} policy over the \textbf{\textit{visuo-proprio}} policy, however such improvement is only 3\% and 7\% with 2/3 and 1/3 training data, respectively.
The same trend is observed with the average success rate as well, with a 34\% and 20\% performance improvement when trained with all data or 2/3 data.
Although multi-modality policies consistently outperform \textbf{\textit{visuo-proprio}} policies, this suggests that more data might be required to unlock the full potential of having multi-modality as inputs.
This issue could be a result of the larger model size, and it could potentially be alleviated with a foundational policy pre-trained with other in-domain data, such as data gathered on different embodiments and/or for different tasks.



\section{Limitations and future works}
Using VHB tape as the elastomer in camera-based tactile sensors is a novel approach with minimized requirements on fabrication equipment and expertise aiming to attract a larger audience in the robotics community to adopt tactile sensors in their pipelines.
Despite its ease of construction and high sensitivity, the hysteresis introduced by the viscosity of the acrylic foam base material used in VHB tapes could lead to slower response or ambiguity in dynamic manipulation tasks.
This issue could be alleviated on the software side, by taking temporal difference images and cross-checking the peripheral vision signal.
Finding a new material with similar ease of construction and less viscosity could be a potential future improvement.

\section{Conclusions}

In this work we presented \name{}, a robust and easy-to-manufacture robot finger that combines tactile, acoustic, and peripheral sensing modalities into a compact form-factor.
Two elastomer formulas, including a novel VHB tape-based formula for ultimate ease of fabrication, are introduced together with their supporting fabrication tools. Our sensor is significantly more durable compared to state-of-the-art alternatives and its compact form factor and multi-modality, naturally render it an ideal choice for large-scale, on-robot policy synthesis research. We then introduce a tactile-diffusion policy framework that combines multi-modal tactile information with vision and proprioception.
Our experiments show that policies trained with multi-modal tactile sensing consistently outperformed state-of-the-art visuomotor policies in contact-rich manipulation.  


\bibliographystyle{IEEEtran}
\bibliography{references}

\end{document}